\documentclass{article}

\usepackage{PRIMEarxiv}

\usepackage[utf8]{inputenc} 
\usepackage[T1]{fontenc}    
\usepackage{hyperref}       
\usepackage{url}            
\usepackage{booktabs}       
\usepackage{amsfonts}       
\usepackage{nicefrac}       
\usepackage{microtype}      
\usepackage{lipsum}
\usepackage{fancyhdr}       
\usepackage{graphicx}       
\graphicspath{{media/}}     
\usepackage{cite}
\usepackage{amsmath,amssymb,amsfonts}
\usepackage{algorithmic}
\usepackage{graphicx}
\usepackage{textcomp}
\usepackage{caption}
\usepackage{url}
\title{UnSegGNet: Unsupervised Image Segmentation using Graph Neural Networks}

\author{
  Kovvuri Sai Gopal Reddy, Bodduluri Saran, A. Mudit Adityaja, Saurabh J. Shigwan, and Nitin Kumar\\
  Shiv Nadar Intitute of Eminence \\
  Delhi-NCR, IN\\
  \texttt{\{kr521,bs404,ma646,saurabh.shigwan,nitin.kumar\}@snu.edu.in} \\
}

\begin{document}


\maketitle

\begin{abstract}
Image segmentation, the process of partitioning an image into meaningful regions, plays a pivotal role in computer vision and medical imaging applications. Unsupervised segmentation, particularly in the absence of labeled data, remains a challenging task due to the inter-class similarity and variations in intensity and resolution. 
In this study, we extract high-level features of the input image using pretrained vision transformer. Subsequently, the proposed method leverages the underlying graph structures of the images, seeking to discover and delineate meaningful boundaries using graph neural networks and modularity based optimization criteria without relying on pre-labeled training data. Experimental results on benchmark datasets demonstrate the effectiveness and versatility of the proposed approach, showcasing competitive performance compared to the state-of-the-art unsupervised segmentation methods. This research contributes to the broader field of unsupervised medical imaging and computer vision by presenting an innovative methodology for image segmentation that aligns with real-world challenges. The proposed method holds promise for diverse applications, including medical imaging, remote sensing, and object recognition, where labeled data may be scarce or unavailable. The github repository of the code is available on \url{https://github.com/ksgr5566/unseggnet}
\end{abstract}

\begin{figure*}
    \centering
    \includegraphics[width=1.1\linewidth,trim={0 0cm 0cm 0},clip]{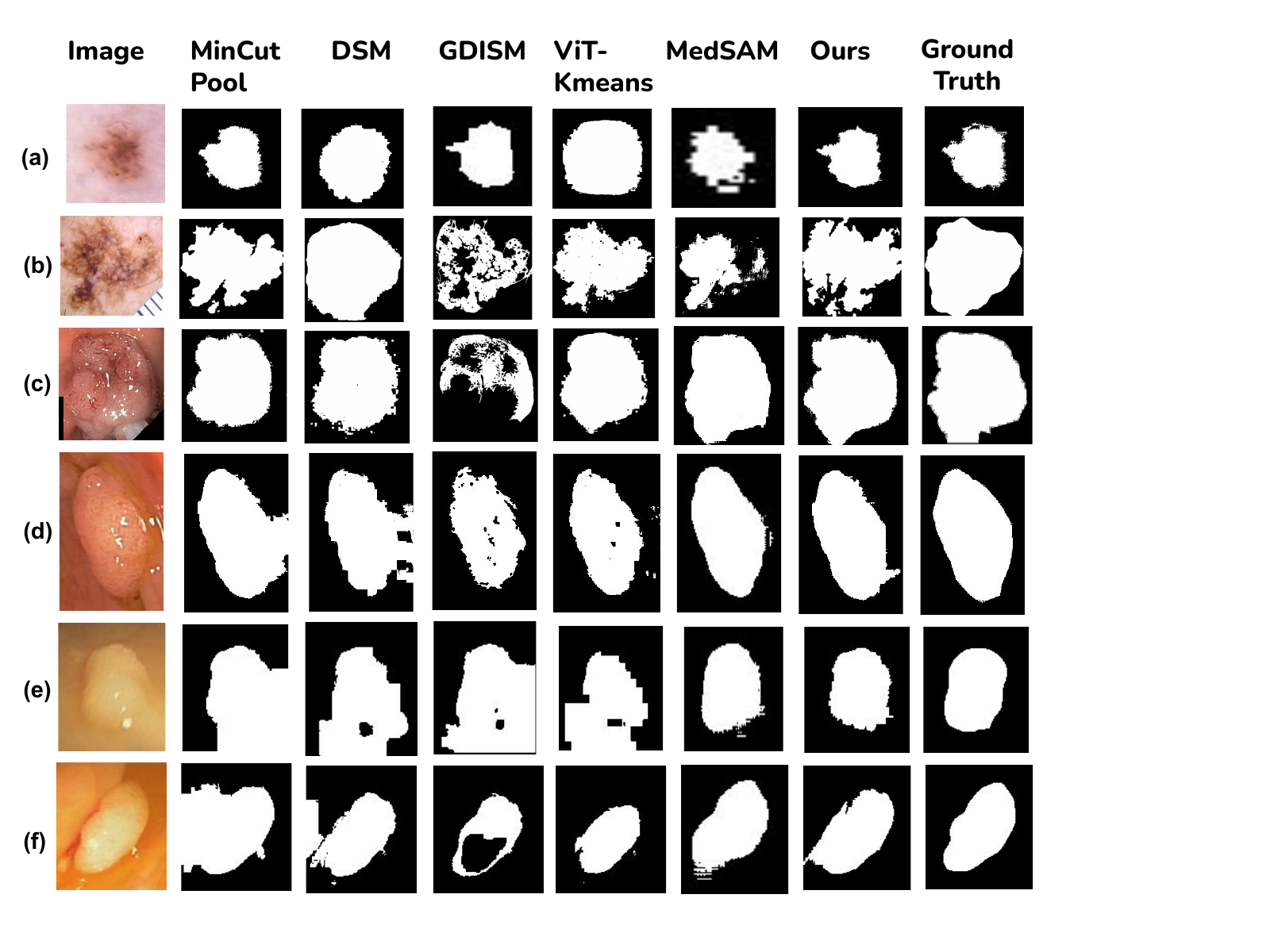}
    \caption{Segmentation results on  (a)-(b) ISIC-2018,  (c)-(d) KVASIR, (e)-(f) CVC-ClinicDB sample images}
    \label{fig:enter-label}
\end{figure*}
\section{Introduction}
\label{sec:introduction}
Segmentation of abnormal regions in medical images is a critical task in medical image analysis, as it facilitates the identification and delineation of pathological areas for treatment and diagnostic purposes. Various methods have been developed and deployed to address this challenge, employing advanced techniques from computer vision and machine learning.

Utilizing annotated datasets, supervised learning methods such as support vector machines~\cite{steinwart2008support}, random forests~\cite{breiman2001random}, or traditional convolutional neural networks (CNNs)~\cite{krizhevsky2012imagenet}, can be trained to segment abnormal regions. In the last decade, CNNs have been widely used for image segmentation tasks.  Popular architectures based on CNNs include UNet\cite{ronneberger2015u}, SegNet~\cite{badrinarayanan2017segnet}, and DeepLab~\cite{chen2017deeplab}, among others. These models incorporate large number of layers to learn hierarchical features/representations from the training data.
These supervised approaches have demonstrated considerable success, although their reliance on annotated datasets poses inherent challenges. As, constructing large-scale annotated datasets is resource-intensive, time-consuming, and often infeasible due to privacy constraints and the variability inherent in medical images. Moreover, the supervised models ~\cite{wang2022medical} trained on a specific dataset may struggle to generalize to new or diverse datasets with variations in imaging protocols, equipment, and patient populations. This lack of generalization can limit the broader applicability of the segmentation model.
One such attempt towards generalization is
MedSAM~\cite{ma2024segment}, the Segment-Anything Model on medical images, which uses vision transformer (ViT)-based image encoder and decoder architecture for image segmentation. It employs pre-trained SAM~\cite{kirillov2023segment} model with the ViT-base~\cite{caron2021emerging} model for fully supervised training over 1 million image-mask pairs spanning across 10 imaging-modalities and over 30 cancer types. One of the objective of MedSAM is to provide a universal model for medical image segmentation which can accommodate a wide range of variations in anatomical structures, imaging and pathological conditions through extensive training over a well curated diverse and big medical image segmentation datasets. But it suffers from modality imbalance issue where certain structures or pathologies may be underrepresented in the dataset~\cite{ma2024segment}.
%
%
Due to this, the segmentation model may bias towards the majority class, leading to suboptimal performance for minority classes.
Also, in supervised segmentation, there exist the problems of inter-observer as well as intra-observer~\cite{schmidt2023probabilistic}\cite{brochez2002inter} variability in the annotations of medical images where in the later case, a single expert may provide different annotations at different times. These variations can introduce challenges in creating consistent and reliable ground truth data which is critical in health care domain.

In response to these limitations, unsupervised medical image segmentation~\cite{huang2004watershed}\cite{aganj2018unsupervised} has emerged as a promising approach, aiming to overcome the dependence on annotated data and alleviating the burden of manual labelling. Unsupervised segmentation techniques operate without explicit annotations, leveraging intrinsic patterns and structures within the images themselves. By exploring the inherent information present in medical images, these methods seek to autonomously identify and delineate relevant regions of interest, offering a potential solution to the growing demand of efficient segmentation methodologies. These methods can be broadly classified into two categories: (i) traditional methods (ii) deep learning based methods. Using tradition unsupervised region based segmentation methods~\cite{roerdink2000watershed}\cite{huang2004watershed}, an image is divided into regions based on similar properties, such as color, intensity, or texture. Subsequently the abnormal regions are distinguished based on intensity differences from the surrounding normal regions. While unsupervised seed based region growing methods~\cite{adams1994seeded}\cite{fan2005seeded} typically start with a random seed points and progressively grow regions based on predefined criteria, such as intensity homogeneity. Level set methods\cite{wang2010efficient}\cite{kass1988snakes} formulate image segmentation as an energy minimization problem and evolve contours to separate regions based on intensity differences. Clustering based segmentation methods such as k-means~\cite{jain1999data} and mean-shift~\cite{cheng1995mean} groups pixels based on color or intensity values and color or texture similarities respectively. Another category of clustering based segmentation is superpixel segmentation which groups pixels into perceptually meaningful atomic regions, known as superpixels. These regions are more homogeneous than individual pixels and often serve as a preprocessing step for segmentation. SLIC (Simple Linear Iterative Clustering)~\cite{achanta2012slic} and Quick Shift~\cite{vedaldi2008quick} are examples of superpixel based segmentation methods. Dictionary learning methods~\cite{yang2014unsupervised} represent image patches sparsely using a learned dictionary, which can help segment structures with distinctive patterns. Another widely popular approach to partition any image is to use the graph as the anatomical representation of the image where the pixel values-based nodes and visual similarity-based edges’ weights [5] can be deployed to perform unsupervised image segmentation~\cite{shi2000normalized}\cite{ melas2022deep}\cite{bianchi2020spectral}\cite{trombini2023goal}. One of the popular instance of the above approach was proposed by~\cite{shi2000normalized} having roots in spectral graph theory. \cite{shi2000normalized} treat the image segmentation  as a graph partitioning problem and proposed normalized cut to segment the graphs into sub-graphs. Normalized cut measures both the total dissimilarity between the different sub-graphs as well as the total similarity within the sub-graphs. It was shown that the solution to the generalized eigenvalue problem~\cite{ghojogh2019eigenvalue}  should be used to optimize the criterion. Specifically, the second smallest eigenvector i.e. Fiedler vector bi-partitions the image into meaningful regions avoiding the trivial isolated regions.
Deep learning architectures such as Autoencoders~\cite{baur2021autoencoders} and Generative Adversarial Networks (GANs)~\cite{zhang2020unsupervised} have been used for unsupervised feature learning and image segmentation by encoding and decoding image representations. Self-organizing maps (SOM)~\cite{kohonen2013essentials} use neural networks to map high-dimensional image data to a lower-dimensional grid, facilitating clustering and segmentation. Recently, unsupervised deep learning based variants~\cite{bianchi2020spectral},\cite{melas2022deep} of normalized-cut have been proposed for image segmentation. \cite{bianchi2020spectral} have been observed to be less effective in clustering the underlying graph structure as shown by~\cite{tsitsulin2023graph}.\cite{melas2022deep} takes significant time to compute the Fiedler vector which is required for segmenting the regions in the image. In contrast to the aforementioned normalized cut variants, our method uses modularity based criteria which results in better performance. We also use the shallow variant of Graph Neural Networks (GNN) to further refine the last-layer features of Vision Transformers (ViT)~\cite{dosovitskiy2020image}. GNNs are particularly useful for tasks involving graph-structured data hence, they are applied to a spatial graph representing the relationships among non-overlapping image patches of the input image. These
relationships are captured by the adjacency matrix or an edge list~\cite{van2023graph}. In our case, we use a special type of normalized adjacency matrix known as modulartiy matrix~\cite{newman2006modularity} to extract powerful dense features for image segmentation. \\
Key components of GNNs are: (i) Neighborhood aggregation: GNNs perform neighborhood aggregation by considering the features and connections of neighboring nodes to update the representation of a given node. In the context of image data, this can involve considering the pixel values of adjacent image pixels or patches. (ii) Message passing: involves passing of information between nodes in multiple iterations or layers. Each layer refines the node representations by incorporating information from the local neighborhood. In the proposed method, nodes are characterized by the image patches which exchange information based on their spatial relationships captured using correlations among their respective features. This helps the model capture context and relationships within the image.  (iii) Graph pooling: GNNs perform graph pooling operations to down-sample the graph structure of the input image. Graph pooling is not required in image segmentation as the underlying graph structure remains invariant.


\begin{figure*}[htbp]
    \centering    \includegraphics[width=0.9\linewidth,trim={0 1.25cm 0 0}]{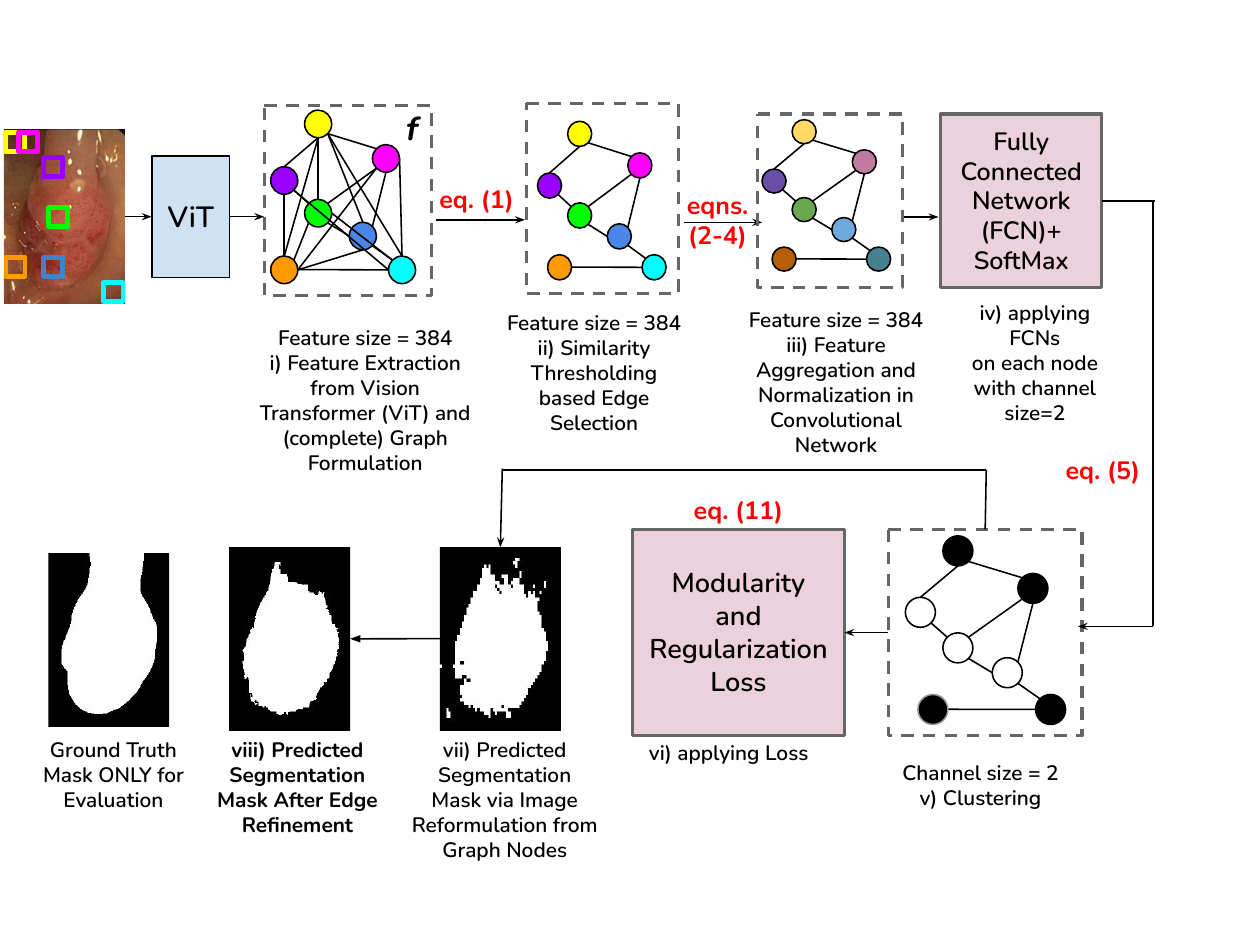}
    \caption{UnSegGNet Pipeline: we i) extract features $f$ of all (overlapping) image patches using vision transformer (ViT) and formulate a (complete) Graph $G$ (few nodes shown, for illustration, in the same color as image patch windows), ii) then apply similarity (normalized $ff^T$) threshold to select important edges in $G$, iii) aggregate and normalize features in graph convolutional network (GCN), darker node colors represent aggregation, iv) apply a fully connected network (FCN) to finally obtain node level clusters. vi) The modularity and regularization based loss is finally used to train the model. vii-viii) At inference, edge refinement is used over the predicted mask.}
    \label{fig:method}
\end{figure*}
\subsection{Contribution of this work}
Unlike its counterparts~\cite{melas2022deep}\cite{wang2023tokencut} which use expensive spectral decomposition to compute eigen-vectors,  the proposed method, UnSegGNet uses modularity matrix~\cite{newman2006modularity} by considering the adjacency of the image patches in the spatial domain followed by the shallow Graph Neural Network (GNN)~\cite{maurya2021improving}. The existing literature on unsupervised image segmentation clearly demonstrates the lack of studies in utilizing the generic pre-trained visual features to segment the complex abnormal regions of medical image datasets. Our method addresses these aforementioned issues.
 The main contributions of the article are: 
\begin{itemize}
    \item We propose \textit{cross-field} unsupervised segmentation framework. The proposed method leverages the extracted features from the pre-trained Vision Transformer~\cite{dosovitskiy2020image} (on ImageNet~\cite{deng2009imagenet}) from vision field and achieves significant performance gain over medical-image datasets by exploiting the underlying graph structure of the input images.
       
    \item We introduce for the first time a modularity~\cite{newman2006modularity} based clustering and loss function, to segment images.


    \item We find that the use of SILU and SELU ~\cite{ramachandran2017searching}~\cite{elfwing2018sigmoid} activation functions in the  Graph Convolution Network outperform the other SOTA methods in medical images and computer vision datasets respectively. 

\end{itemize}

\section{Materials and Methods}
\label{sec:materials_and_methods}

Fig.~\ref{fig:method} depicts an overview of the UnSegGNet 
 architecture. In the pre-processing step, images
are resized to $224 \times 224$ as required by DINO-ViT model~\cite{caron2021emerging}. It has three components:
(i) Pretrained Network: used for extracting high-level features. We use DINO-ViT which produces potentially powerful features trained on Imagenet-21k~\cite{kolesnikov2020big} dataset  (ii) Graph Neural Network~\cite{scarselli2008graph}: used for clustering image patches of the input image based on extracted features from DINO-ViT. In our case, GNN employs efficient  graph convolution, activation function and graph pooling to realize effective clustering of the image patches (iii) Loss Function~\cite{tsitsulin2023graph} : used for optimization of the segmentation based on clustering criteria.
\subsection{Pretrained Network}
\label{sec:pretrain}
As shown in Fig.~\ref{fig:method} (i), we use the features $f$ extracted from vision transformer~\cite{caron2021emerging} which is a deep learning model architecture that extends the transformer architecture, originally designed for natural language processing tasks to computer vision tasks. ViTs handle images of variable sizes by dividing them into fixed-size patches, making it more scalable than traditional convolutional neural networks (CNNs) in terms of both computational efficiency and performance. It also effectively captures global contextual information over image data using  self-attention mechanism~\cite{ramachandran2019stand}\cite{zhao2020exploring}. ViT is not as sensitive to data augmentation as CNNs~\cite{paul2022vision}, hence can be trained on smaller data-sets as well.

In our method, each input image of size $s\times t$ and $d$ channels is processed through a pretrained small vision transformer (ViT)~\cite{caron2021emerging} without fine-tuning to produce deep features. The transformer partitions the image into $st/p^2$ patches, with $p \times p$ being the patch size of the vision transformer. Each patch's internal representation is derived from the key layer features of the final transformer block, recognized for superior performance across diverse tasks. The resulting outputs are feature vectors $f_{(st/p^2) \times C_{in}}$, where $C_{in}$ signifies the token embedding dimension of the corresponding image patch. This is followed by the formation of an input image-specific graph $G$ using the neighborhood relationships through correlation by (\ref{eq:thresh_corr})  whose nodes characterize the image patches. Different patch windows and their corresponding nodes are shown in common colors in Fig.~\ref{fig:method}'s input image and Fig.~\ref{fig:method} (i).

\subsection{Graph Neural Network}
Graph Neural Network (GNN)~\cite{scarselli2008graph} is a type of neural network explicitly designed for processing graph-structured data.
We use a specific type of GNN called  Graph Convolutional Network (GCN) which have graph convolutional layer as its core component. Graph convolutional layer  enables localized spectral filtering~\cite{kipf2016semi} to capture relationships between nodes characterizing the image patches.
These neighborhood relationships  are measured using the thresholded correlation between the corresponding node features $f$ such that,
\begin{equation}
\label{eq:thresh_corr}
A= \biggr(ff^T > \tau\biggr) \in R^{\frac{st}{p^2}\times \frac{st}{p^2}}
\end{equation}
where elements of adjacency matrix $A$ are $A_{ij} \in \{0,1\}$ and $\tau$ is a user defined parameter which is tuned specific to the dataset. The illustrative result of applying equation (\ref{eq:thresh_corr}) on the node features $f$ of $G$ is shown in Fig.~\ref{fig:method} (ii).
Information from neighboring nodes is aggregated as:
\begin{equation}
h_i^{(l)} = \sigma\left( \sum_{j \in \mathcal{N}_i} \frac{1}{\sqrt{d_i d_j}} h_j^{(l-1)}W_j^{(l)} + b_j^{(l)} \right)
\label{eq:gcn_sum}
\end{equation}
where $\sigma$ is the activation function, $\mathcal{N}_i$ represents the set of neighboring nodes of node $i$, and $d_i$ is the degree of node $i$. Here, $h_j^{l}$ denotes the $j^{th}$ node representation after $l^{th}$ pass with learnable weighting parameter $W_j^{l}$ and bias term $b_j^{l}$. Specifically, $h_j^{0} \in \mathcal{R}^{C_{in}}$ is the feature of the $j^{th}$ patch and $h_j^{l} \in \mathcal{R}^{C_{hidden}}$  is $l^{th}$ intermediate layer feature of the $j^{th}$ patch.
Furthermore, $W^{0} \in \mathcal{R}^{C_{in}\times C_{hidden}}$ and $W^{k} \in \mathcal{R}^{C_{hidden}\times C_{hidden}}$ are initialized with Glorot~\cite{glorot2010understanding} uniform distribution with $C_{hidden}$ as the number of hidden layers in GCN. Here, we have $C_{hidden} = 64$ and $C_{in}=384$.
Summation over neighborhood in equation (\ref{eq:gcn_sum}) can be expressed as normalized adjacency matrix $\hat{A}$,
\begin{equation}
\hat{A} = D^{-\frac{1}{2}} A D^{-\frac{1}{2}}
\end{equation}

where $D$ and $A$ are the degree matrix (with $D_{ii}=d_i$) and adjacency matrix respectively.
Hence, equation (\ref{eq:gcn_sum}) can be represented in matrix form:
\begin{equation}
H^{(l)} = \sigma(\hat{A}H^{(l-1)}W^{(l)})
\label{eq:GCN}
\end{equation}
where $H^{(0)}=f$.
The equation (\ref{eq:GCN}) represents multi-pass GCN layer as mentioned in~\cite{kipf2016semi}. The illustrative result of applying equations (\ref{eq:gcn_sum}-\ref{eq:GCN}) is shown in  Fig.~\ref{fig:method} (iii). Note that the darker colors in Fig.~\ref{fig:method} (iii) compared to Fig.~\ref{fig:method} (ii), represent feature aggregation.

Output of the final $L^{th}$ layer $H^{(L)}$, is passed to FCN layer and softmax to get final cluster assigment $C \in[0, 1] ^{n\times k}$. Here, $n,k$ are the number of nodes and clusters respectively as mentioned in equation (\ref{eq:cluster_C}) (see Fig.~\ref{fig:method} (iv), where a black node represents 0 class, and a white node represents 1 class).
\begin{equation}
 C = softmax(MLP(GCN(\hat  {A},X))) 
 \label{eq:cluster_C}
\end{equation}

\subsection{Loss Function}
The loss function is based on the modularity matrix $B$ which is defined on an undirected graph $\mathcal{G}= (\mathcal{V},\mathcal{E})$ where $\mathcal{V}=(v_1,\ldots,v_n),$ is the set of $n$ nodes and edges $\mathcal{E} \subseteq \mathcal{V} \times \mathcal{V}$. If $A$ is the adjacency matrix of the undirected graph $\mathcal{G}$ such that $A_{ij}=1 \hspace{3pt}$ when $\hspace{3pt}\{v_i,v_j\} \in \mathcal{E}$, else $A_{ij}=0$ then
\begin{equation}
\label{eq:mod_matrix}
B= A - \frac{dd^T}{2m}
\end{equation}
 where $d$ is the degree vector of $\mathcal{G}$ and $m=|\mathcal{E}|$ i.e. the number of edges in $\mathcal{G}$. Here, the modularity matrix $B$ reflects the differences between the observed edges in the graph and the expected edges in a random graph with the same degree sequence. Hence, the positive values in the modularity matrix indicate higher density of edges within the partitioned subgraph (cluster), contributing to a higher modularity score. Subsequently, modularity of any partition $C$ of the graph can be expressed as:
\begin{equation} 
\label{eq:modularity}
Q = \frac{1}{2m}\sum_{ij}\bigg[A_{ij}-\frac{d_i d_j}{2m}\bigg]\delta(c_i,c_j)
\end{equation}
where $c_i$ represents the cluster to which node $i$ belongs,
$\delta$ is the Kronecker delta function defined as:
\begin{equation}
\delta(a,b) =
\begin{cases}
 1, \hspace{5pt} if \hspace{5pt} a = b \\
 0,  \hspace{5pt} if \hspace{5pt} a \neq b
 \end{cases}
\end{equation}
 and the summation is over all the pairs of nodes.\\
As the higher modularity score signifies better quality of the partitioning of graph $\mathcal{G}$, maximizing the modularity $Q$ reflects better clustering. Also, maximizing equation (\ref{eq:modularity}) is proven to be NP-hard~\cite{tsitsulin2023graph} so its relaxed version  (\ref{eq:rel_bar_Q}) is considered for optimization.
\begin{equation}
\label{eq:rel_bar_Q}
\Bar{Q}= \frac{1}{2m}Tr(C^TBC)
\end{equation}
 Where the cluster assignment matrix $C \in\mathbb{R} ^{n\times k}$ ($n,k$ are number of nodes and clusters respectively) is obtained using equation (\ref{eq:cluster_C}).
 $Tr(\cdot)$ is the trace operator which computes the sum of the diagonal entries of the square matrix. We cluster the resulting (patch) features from section \ref{sec:pretrain} by optimizing the loss function $\mathcal{L}$ in equation (\ref{eq:loss_fun}) which guarantees to produce non-trivial cluster assignment matrix $C$~\cite{tsitsulin2023graph}. 
 \begin{equation}
 \label{eq:loss_fun}
 \mathcal{L} = -\frac{1}{2m}Tr(C^TBC) + \frac{\sqrt{k}}{n}\bigg\lVert \sum_{i=1}^{n} C_i \bigg\rVert_F - 1  
 \end{equation}
Where $C_i\in [0,1]^{1\times k}$ is the soft cluster assignments of the $i^{th}$ node and $\lVert.\rVert_F$ is the Frobenius norm.

\section{Experiments and Results}

 \subsection{Experimental Details}
 
 We use DINO-ViT small model with patch size
 $p = 8$, which gives node features of size $C_{in} = 384$ for our UnSegGNet framework. It uses two layered GCN ($L=2$) to aggregate the feature which are subsequently projected from dimension $384$ to $64$, and $64$ to $k$ respectively using FCN for the binary segmentation ($k=2$). UnSegGNet is optimized separately for each image using loss function $\mathcal{L}$. The initial learning rate is set to $10^{-3}$ with the decay of $10^{-2}$. For optimizer, we use ADAM~\cite{kingma2014adam} with $100$ epochs.
 We observe  using $ 0.4 \le \tau \le 0.6$ in the model provide better accuracy and fast convergence over different datasets.
 We also experimented with different activation functions like SiLU~\cite{elfwing2018sigmoid}, SeLU~\cite{ramachandran2017searching}, GELU~\cite{hendrycks2016gaussian} and ReLU~\cite{glorot2011deep}, and found SiLU and SeLU to be most effective for our work.
 
\subsection{Datasets}
\subsubsection{CVC-ClinicDB~\cite{bernal2015wm}}
is a well-known open-access dataset for colonoscopy research. It was introduced as part of the Computer Vision Center (CVC) Colon Image Database project. It contains 612 images with a resolution of 384×288 from 31 colonoscopy sequences acquired during routine colonoscopies.
\subsubsection{KVASIR~\cite{jha2020kvasir}} includes images from the gastrointestinal tract, obtained through various endoscopic procedures such as gastroscopy and colonoscopy. The dataset contains 8,000 endoscopic images, with 1,000 image examples per class.  Images in the KVASIR dataset are typically labeled with 8 different classes related to gastrointestinal conditions, lesions, or abnormalities. 
\subsubsection{ISIC-2018~\cite{codella2019skin}}
The ISIC (International Skin Imaging Collaboration) 2018 is a dataset and challenge focused on dermatology and skin cancer detection. The dataset is designed to facilitate research in the development of computer algorithms for the automatic diagnosis of skin lesions, including the identification of melanoma. The dataset includes clinical images, dermoscopic images, and images obtained with confocal microscopy.
\subsubsection{ECSSD~\cite{shi2015hierarchical}}  ECSSD (Extended Complex Scene Stereo Dataset) is a dataset commonly used  to identify the most visually significant regions within an image. It has been designed to evaluate the performance of algorithms on complex and natural scenes. The dataset contains 1000 RGB images captured in complex and diverse scenes along with the annotated pixel-wise ground-truth masks indicating the saliency regions created as an average of the labeling of five human participants.
\subsubsection{DUTS~\cite{wang2017learning}} 
The DUTS (Deep Unsupervised Training for Salient Object Detection) dataset is widely used in the field of computer vision for the evaluation of salient object detection algorithms. It contains 10,553 training images and 5,019 test images. 
The dataset consists of images with diverse and complex scenes and each image is annotated with pixel-wise ground truth mask which indicates the salient object within the scene. We use only test dataset for evaluation. 
\subsubsection{CUB~\cite{wah2011caltech}} CUB (Caltech-UCSD Birds) dataset is a widely used dataset in the field of computer vision, particularly for the classification of bird species. It contains images of 200 bird species, with a total of around 11,788 images. Each image is associated with a specific bird species. The dataset provides detailed annotations
for bird parts such as head, body, and tail .

\subsection{Results}
\label{sec:exp_det}
We compare UnSegGNet with other SOTA methods on the following publicly available benchmarks: KVASIR~\cite{jha2020kvasir}, CVC-ClinicDB~\cite{bernal2015wm}, ISIC-2018~\cite{codella2019skin}, ECSSD~\cite{shi2015hierarchical}, DUTS~\cite{wang2017learning}, CUB~\cite{wah2011caltech}. 
KVASIR, CVC-ClinicDB, and ISIC-2018 are medical image datasets while ECSSD, DUTS and CUB are non-medical general purpose vision datasets. Initially, we evaluate the performance of our method on the ECSSD, DUTS, and CUB datasets, in accordance with previous assessments of state-of-the-art methods ~\cite{bianchi2020spectral}~\cite{melas2022deep}~\cite{aflalo2023deepcut}. Our method yields better results compared to unsupervised SOTA methods, as shown in the table \ref{table:tab2}. 

We utilize the following methods for comparison with medical image datasets: MedSAM\cite{ma2024segment}, MinCutPool\cite{bianchi2020spectral}, GDISM\cite{trombini2023goal}, DSM\cite{melas2022deep} and ViT-Kmeans. As shown in the table~\ref{table:tab1}, UnSegGNet outperforms unsupervised SOTA methods. UnSegGNet produces significantly better scores on ISIC-2018 compared to MedSAM. MedSAM produces best results on KVASIR and CVC-ClinicDB due to being trained on very large polyp datasets while it produces inferior results to UnSegGNet on ISIC-2018 due to the modality-imbalance issue.\\ 
We also evaluated UnSegGNet on the following activation functions: SiLU~\cite{elfwing2018sigmoid}, SeLU~\cite{ramachandran2017searching}, GELU~\cite{hendrycks2016gaussian} and ReLU~\cite{glorot2011deep}. Results on medical and vision datasets using these activation functions are shown in  Table~\ref{table:tab3}. UnSegGNet with SiLU activation function produces best results on all the three medical image datasets while the use of SeLU activation function provide best results on other benchmarks compared to SOTA unsupervised methods.

%
\begin{table*} [t]
\centering
\begin{tabular}{l|cccccc}
 \textbf{Datasets}
& \textbf{UnSegGNet (Ours)} & \textbf{MedSAM}~\cite{ma2024segment} & \textbf{MinCutPool~\cite{bianchi2020spectral}} & \textbf{GDISM\cite{trombini2023goal}} & \textbf{DSM}~\cite{melas2022deep} & \textbf{ViT-Kmeans} \\ 
\midrule \hline
KVASIR & 74 & 76.72 & 73.50 & 59.40 & 58.80 & 66.10 \\
ISIC-2018 & 73.94 & 61.36 & 72.31 & 52.53 & 72.20 & 68.60  \\
CVC-ClinicDB & 64 & 71.53 & 62.10 & 59.22 & 60.48 & 62.80  \\
\end{tabular}
\caption{Average mIOU scores for medical image data}
\label{table:tab1}
\end{table*}
We consider binarized value in the segmentation results evaluated using Mean Intersection over Union (mIOU) scores.  mIoU is the average of the IoU scores across all $N$ classes and is computed as:  
\begin{equation}
\label{eq:miou}
    mIOU = \frac{1}{N}\sum_{i=1}^{N}IoU_i
\end{equation}
where IoU for class $i$ is:
\begin{equation}
\label{eq:iou}
    IOU_i = \frac{TP_i}{TP_i + FP_i + FN_i}
\end{equation}
$TP_i$ is the true positive (the number of pixels that are correctly classified as class $i$), $FP_i$ is the false positive (the number of pixels that are incorrectly classified as class $i$) and $FN_i$ is the false negative (the number of pixels that are incorrectly classified as different class other than class $i$. mIoU provides a comprehensive evaluation of the model's segmentation performance by considering the overlap between predicted and ground truth masks for each class. Higher mIoU values indicate better segmentation accuracy.\\

\textbf{Results on Medical Image Datasets.}
As observed from Table~\ref{table:tab1} , UnsegGNet clearly outperforms other unsupervised methods. UnSegGNet produces comparative results to MedSAM on KVASIR~\cite{jha2020kvasir} and ISIC-2018~\cite{codella2019skin}, which has been fine-tuned on large pool of medical image datasets including large colonoscopy datasets as well. One of the issue with MedSAM seems to be the problem of modality-imbalance. Notably, it used large polyp and small dermoscopy image-mask pairs in the training phase, which would have caused comparatively better performance on KVASIR and CVC-ClinicDB datasets and significantly lower scores on ISIC-2018. \\
\begin{figure*}
    \centering
    \includegraphics[width=1.0\linewidth,trim={0 0cm 0cm 0},clip]{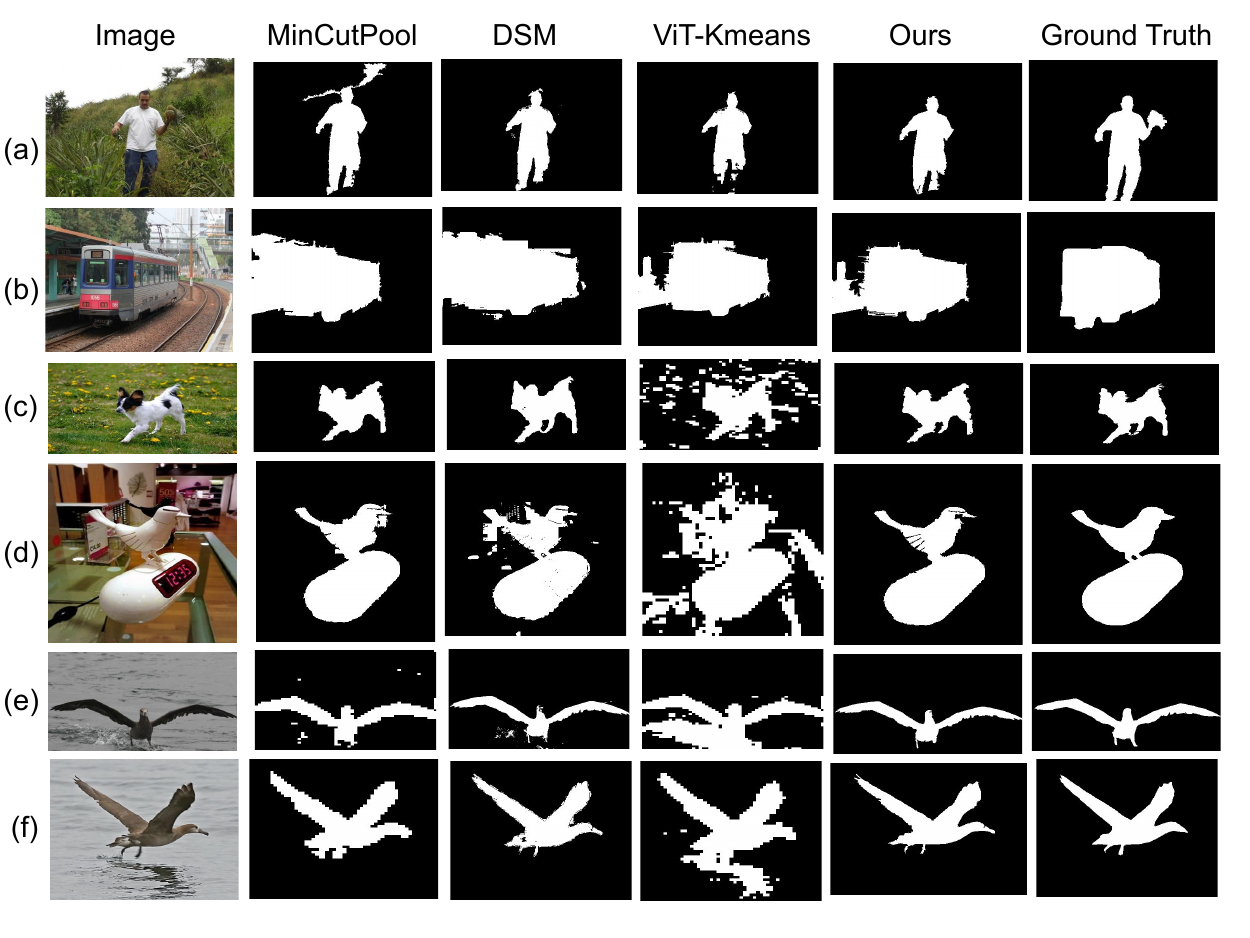}
    \caption{Segmentation results on  (a)-(b) ECSSD,  (c)-(d) DUTS, (e)-(f) CUB sample images}
    \label{fig:cv_results}
\end{figure*}
\begin{table}[t]
\centering
\setlength{\tabcolsep}{2pt}
\begin{tabular}{l|cccc}
\textbf{Datasets}
& \textbf{Ours}  & \textbf{MinCutPool\cite{bianchi2020spectral}} & \textbf{DSM}\cite{melas2022deep} & \textbf{ViT-Kmeans} \\ 
\midrule \hline
ECSSD & 75.57 & 74.6 & 73.3 & 65.76\\
DUTS & 61.04 & 59.5 & 51.4 & 42.78\\
CUB & 78.41 & 78.2 & 76.9 & 51.86\\
\end{tabular}
\caption{Average mIOU scores for computer vision datasets}
\label{table:tab2}
\end{table}
\textbf{Results on Other Datasets.}
We show results on publicly available ECSSD, CUB and DUTs vision datasets by comparing it with MinCutPool~\cite{bianchi2020spectral}, DSM~\cite{melas2022deep} and ViT-Kmeans. Table II summarizes the scores. We find that UnSegGNet performs fairly better than the other methods. Similar to UnSegGNet, these methods deploy ViT to extract high-level features. ViT-Kmeans directly employ k-means algorithm~\cite{kanungo2002efficient}  on the features extracted from ViT. We observe that there exist significant differences between the scores of UnSegGNet and ViT-Kmeans on all the datasets, which shows that high-level features extracted from ViT alone are not enough for effective segmentation. We also validate that our modularity based UnSegGNet framework gives superior performance on all the benchmarks than ~\cite{bianchi2020spectral} and ~\cite{melas2022deep} which use graph-cut based spectral decomposition in their respective deep learning based architectures.
\begin{table} [h]
\centering
\begin{tabular}{l|ccccc}
 \textbf{Datasets}
& \textbf{SiLU} & \textbf{SeLU} & \textbf{GELU} &  \textbf{ReLU}  \\ 
\midrule \hline
KVASIR & 74 & 45.043 & 68.94 &  45.04 \\
ISIC-2018 & 73.94 & 63.93 & 73.74  & 63.93 \\
CVC-ClinicDB & 64 & 34.90 & 60.05 & 34.90 \\
ECSSD & 57.89 & 75.57 & 57.76 & 70.26\\
DUTS & 55.73 & 61.04 & 35.66 & 60.20\\
CUB & 77.91 & 78.41 & 76.23 & 70.21\\
\end{tabular}
\caption{Average mIOU scores of different activation functions for medical image data}
\label{table:tab3}
\end{table}

\section{Discussion and Conclusion}
\label{sec:guidelines}
 The proposed approach consists of two parts: pretrained network, GNN and its optimization. We use DINO-ViT small as a pretrained network. DINO-ViT is a widely used pretrained feature extractor as it provides robust features which are also adaptable to different data modalities~\cite{paul2022vision}. 
However, ViT features produce coarse segmentation~\cite{melas2022deep} which we verify by applying k-means on these features (ViT-Kmeans). Hence ViT features alone are not sufficient to produce good quality segmentation results as shown in Fig.\ref{fig:enter-label} and Fig.\ref{fig:cv_results}.

Though \cite{ma2024segment} proposes generalized model $MedSAM$ for universal medical image segmentation, it is prone to modality imbalance problem as mentioned in section~\ref{sec:exp_det} . Also, if images from all the modalities are equal in proportion still it  does not always guarantee to provide fair accuracies. It is also difficult to update such model due to high computation cost.  

MinCutPool~\cite{bianchi2020spectral} proposes a novel pooling policy based on the relaxation of the normalized min-cut~\cite{shi2000normalized} problem. This approach uses spatially localized graph convolutions in the graph neural networks and hence avoids the computation of expensive spectral decomposition. DSM~\cite{melas2022deep} encodes similarities based on the correlation of ViT features and color information in the graph adjacency matrix $A$. Subsequently, the spectral decomposition is performed on $A$ to segment the input image. However, spectral decomposition is computationally expensive.

GDISM~\cite{trombini2023goal} is a parametric and graph based unsupervised method of image segmentation which generates image segments based on user-defined application specific goals. It initializes a set of points called seeds and subsequently identifies homogeneous and non-granular regions by minimizing an ad hoc energy function based on the criteria of graph based methods and Markov random fields. We observe that underlying methodologies in GDISM effectively capture the local contextual informatin which results in limited mIoU as is evident from the results of table  \ref{table:tab1}.

Although DINO-ViT produces useful global features, still this results in coarse segmentation of images. Our proposed modularity based GNN framework leverages these feature to efficiently capture both local as well as global context of images from different modalities. Apart from this, UnSegGNet is a shallow network which requires less time and efforts to segment the input images. Through extensive experimentation across six
real-world datasets of different sizes, we have demonstrated the superior performance of UnSegGNet.

\section*{Declarations}
\textbf{Conflict of interest}
The authors declare that they have no conflict of interest.\\
\textbf{Data availability}
All datasets used for our experiments are freely available.

\bibliographystyle{unsrt}  
\bibliography{reference}

\begin{thebibliography}{10}

\bibitem{steinwart2008support}
Ingo Steinwart and Andreas Christmann.
\newblock {\em Support vector machines}.
\newblock Springer Science \& Business Media, 2008.

\bibitem{breiman2001random}
Leo Breiman.
\newblock Random forests.
\newblock {\em Machine learning}, 45:5--32, 2001.

\bibitem{krizhevsky2012imagenet}
Alex Krizhevsky, Ilya Sutskever, and Geoffrey~E Hinton.
\newblock Imagenet classification with deep convolutional neural networks.
\newblock {\em Advances in neural information processing systems}, 25, 2012.

\bibitem{ronneberger2015u}
Olaf Ronneberger, Philipp Fischer, and Thomas Brox.
\newblock U-net: Convolutional networks for biomedical image segmentation.
\newblock In {\em Medical image computing and computer-assisted intervention--MICCAI 2015: 18th international conference, Munich, Germany, October 5-9, 2015, proceedings, part III 18}, pages 234--241. Springer, 2015.

\bibitem{badrinarayanan2017segnet}
Vijay Badrinarayanan, Alex Kendall, and Roberto Cipolla.
\newblock Segnet: A deep convolutional encoder-decoder architecture for image segmentation.
\newblock {\em IEEE transactions on pattern analysis and machine intelligence}, 39(12):2481--2495, 2017.

\bibitem{chen2017deeplab}
Liang-Chieh Chen, George Papandreou, Iasonas Kokkinos, Kevin Murphy, and Alan~L Yuille.
\newblock Deeplab: Semantic image segmentation with deep convolutional nets, atrous convolution, and fully connected crfs.
\newblock {\em IEEE transactions on pattern analysis and machine intelligence}, 40(4):834--848, 2017.

\bibitem{wang2022medical}
Risheng Wang, Tao Lei, Ruixia Cui, Bingtao Zhang, Hongying Meng, and Asoke~K Nandi.
\newblock Medical image segmentation using deep learning: {A} survey.
\newblock {\em IET Image Processing}, 16(5):1243--1267, 2022.

\bibitem{ma2024segment}
Jun Ma, Yuting He, Feifei Li, Lin Han, Chenyu You, and Bo~Wang.
\newblock Segment anything in medical images.
\newblock {\em Nature Communications}, 15(1):654, 2024.

\bibitem{kirillov2023segment}
Alexander Kirillov, Eric Mintun, Nikhila Ravi, Hanzi Mao, Chloe Rolland, Laura Gustafson, Tete Xiao, Spencer Whitehead, Alexander~C Berg, Wan-Yen Lo, et~al.
\newblock Segment anything.
\newblock In {\em Proceedings of the IEEE/CVF International Conference on Computer Vision}, pages 4015--4026, 2023.

\bibitem{caron2021emerging}
Mathilde Caron, Hugo Touvron, Ishan Misra, Herv{\'e} J{\'e}gou, Julien Mairal, Piotr Bojanowski, and Armand Joulin.
\newblock Emerging properties in self-supervised vision transformers.
\newblock In {\em Proceedings of the IEEE/CVF international conference on computer vision}, pages 9650--9660, 2021.

\bibitem{schmidt2023probabilistic}
Arne Schmidt, Pablo Morales-{\'A}lvarez, and Rafael Molina.
\newblock Probabilistic modeling of inter-and intra-observer variability in medical image segmentation.
\newblock In {\em Proceedings of the IEEE/CVF International Conference on Computer Vision}, pages 21097--21106, 2023.

\bibitem{brochez2002inter}
Lieve Brochez, Evelien Verhaeghe, Edouard Grosshans, Eckhart Haneke, G{\'e}rald Pi{\'e}rard, Dirk Ruiter, and Jean-Marie Naeyaert.
\newblock Inter-observer variation in the histopathological diagnosis of clinically suspicious pigmented skin lesions.
\newblock {\em The Journal of Pathology: A Journal of the Pathological Society of Great Britain and Ireland}, 196(4):459--466, 2002.

\bibitem{huang2004watershed}
Yu-Len Huang and Dar-Ren Chen.
\newblock Watershed segmentation for breast tumor in 2-{D} sonography.
\newblock {\em Ultrasound in medicine \& biology}, 30(5):625--632, 2004.

\bibitem{aganj2018unsupervised}
Iman Aganj, Mukesh~G Harisinghani, Ralph Weissleder, and Bruce Fischl.
\newblock Unsupervised {Medical Image Segmentation Based on the Local Center of M}ass.
\newblock {\em Scientific reports}, 8(1):13012, 2018.

\bibitem{roerdink2000watershed}
Jos~BTM Roerdink and Arnold Meijster.
\newblock The {Watershed Transform: Definitions, Algorithms and Parallelization S}trategies.
\newblock {\em Fundamenta informaticae}, 41(1-2):187--228, 2000.

\bibitem{adams1994seeded}
Rolf Adams and Leanne Bischof.
\newblock Seeded region growing.
\newblock {\em IEEE Transactions on pattern analysis and machine intelligence}, 16(6):641--647, 1994.

\bibitem{fan2005seeded}
Jianping Fan, Guihua Zeng, Mathurin Body, and Mohand-Said Hacid.
\newblock Seeded region growing: an extensive and comparative study.
\newblock {\em Pattern recognition letters}, 26(8):1139--1156, 2005.

\bibitem{wang2010efficient}
Xiao-Feng Wang, De-Shuang Huang, and Huan Xu.
\newblock An efficient local chan--vese model for image segmentation.
\newblock {\em Pattern Recognition}, 43(3):603--618, 2010.

\bibitem{kass1988snakes}
Michael Kass, Andrew Witkin, and Demetri Terzopoulos.
\newblock Snakes: Active contour models.
\newblock {\em International journal of computer vision}, 1(4):321--331, 1988.

\bibitem{jain1999data}
Anil~K Jain, M~Narasimha Murty, and Patrick~J Flynn.
\newblock Data clustering: a review.
\newblock {\em ACM computing surveys (CSUR)}, 31(3):264--323, 1999.

\bibitem{cheng1995mean}
Yizong Cheng.
\newblock Mean shift, mode seeking, and clustering.
\newblock {\em IEEE transactions on pattern analysis and machine intelligence}, 17(8):790--799, 1995.

\bibitem{achanta2012slic}
Radhakrishna Achanta, Appu Shaji, Kevin Smith, Aurelien Lucchi, Pascal Fua, and Sabine S{\"u}sstrunk.
\newblock Slic superpixels compared to state-of-the-art superpixel methods.
\newblock {\em IEEE transactions on pattern analysis and machine intelligence}, 34(11):2274--2282, 2012.

\bibitem{vedaldi2008quick}
Andrea Vedaldi and Stefano Soatto.
\newblock Quick {Shift and Kernel Methods for Mode S}eeking.
\newblock In {\em Computer Vision--ECCV 2008: 10th European Conference on Computer Vision, Marseille, France, October 12-18, 2008, Proceedings, Part IV 10}, pages 705--718. Springer, 2008.

\bibitem{yang2014unsupervised}
Shuyuan Yang, Yuan Lv, Yu~Ren, Lixia Yang, and Licheng Jiao.
\newblock Unsupervised images segmentation via incremental dictionary learning based sparse representation.
\newblock {\em Information Sciences}, 269:48--59, 2014.

\bibitem{shi2000normalized}
Jianbo Shi and Jitendra Malik.
\newblock Normalized cuts and image segmentation.
\newblock {\em IEEE Transactions on pattern analysis and machine intelligence}, 22(8):888--905, 2000.

\bibitem{melas2022deep}
Luke Melas-Kyriazi, Christian Rupprecht, Iro Laina, and Andrea Vedaldi.
\newblock Deep spectral methods: A surprisingly strong baseline for unsupervised semantic segmentation and localization.
\newblock In {\em Proceedings of the IEEE/CVF Conference on Computer Vision and Pattern Recognition}, pages 8364--8375, 2022.

\bibitem{bianchi2020spectral}
Filippo~Maria Bianchi, Daniele Grattarola, and Cesare Alippi.
\newblock Spectral clustering with graph neural networks for graph pooling.
\newblock In {\em International conference on machine learning}, pages 874--883. PMLR, 2020.

\bibitem{trombini2023goal}
Marco Trombini, David Solarna, Gabriele Moser, and Silvana Dellepiane.
\newblock A goal-driven unsupervised image segmentation method combining graph-based processing and markov random fields.
\newblock {\em Pattern Recognition}, 134:109082, 2023.

\bibitem{ghojogh2019eigenvalue}
Benyamin Ghojogh, Fakhri Karray, and Mark Crowley.
\newblock Eigenvalue and generalized eigenvalue problems: Tutorial.
\newblock {\em arXiv preprint arXiv:1903.11240}, 2019.

\bibitem{baur2021autoencoders}
Christoph Baur, Stefan Denner, Benedikt Wiestler, Nassir Navab, and Shadi Albarqouni.
\newblock Autoencoders for unsupervised anomaly segmentation in brain mr images: a comparative study.
\newblock {\em Medical Image Analysis}, 69:101952, 2021.

\bibitem{zhang2020unsupervised}
Yue Zhang, Shun Miao, Tommaso Mansi, and Rui Liao.
\newblock Unsupervised x-ray image segmentation with task driven generative adversarial networks.
\newblock {\em Medical image analysis}, 62:101664, 2020.

\bibitem{kohonen2013essentials}
Teuvo Kohonen.
\newblock Essentials of the self-organizing map.
\newblock {\em Neural networks}, 37:52--65, 2013.

\bibitem{tsitsulin2023graph}
Anton Tsitsulin, John Palowitch, Bryan Perozzi, and Emmanuel M{\"u}ller.
\newblock Graph clustering with graph neural networks.
\newblock {\em Journal of Machine Learning Research}, 24(127):1--21, 2023.

\bibitem{dosovitskiy2020image}
Alexey Dosovitskiy, Lucas Beyer, Alexander Kolesnikov, Dirk Weissenborn, Xiaohua Zhai, Thomas Unterthiner, Mostafa Dehghani, Matthias Minderer, Georg Heigold, Sylvain Gelly, et~al.
\newblock An image is worth 16x16 words: Transformers for image recognition at scale.
\newblock {\em arXiv preprint arXiv:2010.11929}, 2020.

\bibitem{van2023graph}
Piet Van~Mieghem.
\newblock {\em Graph spectra for complex networks}.
\newblock Cambridge University Press, 2023.

\bibitem{newman2006modularity}
Mark~EJ Newman.
\newblock Modularity and community structure in networks.
\newblock {\em Proceedings of the national academy of sciences}, 103(23):8577--8582, 2006.

\bibitem{wang2023tokencut}
Yangtao Wang, Xi~Shen, Yuan Yuan, Yuming Du, Maomao Li, Shell~Xu Hu, James~L Crowley, and Dominique Vaufreydaz.
\newblock Tokencut: Segmenting objects in images and videos with self-supervised transformer and normalized cut.
\newblock {\em IEEE transactions on pattern analysis and machine intelligence}, 2023.

\bibitem{maurya2021improving}
Sunil~Kumar Maurya, Xin Liu, and Tsuyoshi Murata.
\newblock Improving graph neural networks with simple architecture design.
\newblock {\em arXiv preprint arXiv:2105.07634}, 2021.

\bibitem{deng2009imagenet}
Jia Deng, Wei Dong, Richard Socher, Li-Jia Li, Kai Li, and Li~Fei-Fei.
\newblock Imagenet: A large-scale hierarchical image database.
\newblock In {\em 2009 IEEE conference on computer vision and pattern recognition}, pages 248--255. Ieee, 2009.

\bibitem{ramachandran2017searching}
Prajit Ramachandran, Barret Zoph, and Quoc~V Le.
\newblock Searching for activation functions.
\newblock {\em arXiv preprint arXiv:1710.05941}, 2017.

\bibitem{elfwing2018sigmoid}
Stefan Elfwing, Eiji Uchibe, and Kenji Doya.
\newblock Sigmoid-weighted linear units for neural network function approximation in reinforcement learning.
\newblock {\em Neural networks}, 107:3--11, 2018.

\bibitem{kolesnikov2020big}
Alexander Kolesnikov, Lucas Beyer, Xiaohua Zhai, Joan Puigcerver, Jessica Yung, Sylvain Gelly, and Neil Houlsby.
\newblock Big transfer (bit): General visual representation learning.
\newblock In {\em Computer Vision--ECCV 2020: 16th European Conference, Glasgow, UK, August 23--28, 2020, Proceedings, Part V 16}, pages 491--507. Springer, 2020.

\bibitem{scarselli2008graph}
Franco Scarselli, Marco Gori, Ah~Chung Tsoi, Markus Hagenbuchner, and Gabriele Monfardini.
\newblock The graph neural network model.
\newblock {\em IEEE transactions on neural networks}, 20(1):61--80, 2008.

\bibitem{ramachandran2019stand}
Prajit Ramachandran, Niki Parmar, Ashish Vaswani, Irwan Bello, Anselm Levskaya, and Jon Shlens.
\newblock Stand-alone self-attention in vision models.
\newblock {\em Advances in neural information processing systems}, 32, 2019.

\bibitem{zhao2020exploring}
Hengshuang Zhao, Jiaya Jia, and Vladlen Koltun.
\newblock Exploring self-attention for image recognition.
\newblock In {\em Proceedings of the IEEE/CVF conference on computer vision and pattern recognition}, pages 10076--10085, 2020.

\bibitem{paul2022vision}
Sayak Paul and Pin-Yu Chen.
\newblock Vision transformers are robust learners.
\newblock In {\em Proceedings of the AAAI conference on Artificial Intelligence}, volume~36, pages 2071--2081, 2022.

\bibitem{kipf2016semi}
Thomas~N Kipf and Max Welling.
\newblock Semi-supervised classification with graph convolutional networks.
\newblock {\em arXiv preprint arXiv:1609.02907}, 2016.

\bibitem{glorot2010understanding}
Xavier Glorot and Yoshua Bengio.
\newblock Understanding the difficulty of training deep feedforward neural networks.
\newblock In {\em Proceedings of the thirteenth international conference on artificial intelligence and statistics}, pages 249--256. JMLR Workshop and Conference Proceedings, 2010.

\bibitem{kingma2014adam}
Diederik~P Kingma and Jimmy Ba.
\newblock Adam: A method for stochastic optimization.
\newblock {\em arXiv preprint arXiv:1412.6980}, 2014.

\bibitem{hendrycks2016gaussian}
Dan Hendrycks and Kevin Gimpel.
\newblock Gaussian error linear units (gelus).
\newblock {\em arXiv preprint arXiv:1606.08415}, 2016.

\bibitem{glorot2011deep}
Xavier Glorot, Antoine Bordes, and Yoshua Bengio.
\newblock Deep sparse rectifier neural networks.
\newblock In {\em Proceedings of the fourteenth international conference on artificial intelligence and statistics}, pages 315--323. JMLR Workshop and Conference Proceedings, 2011.

\bibitem{bernal2015wm}
Jorge Bernal, F~Javier S{\'a}nchez, Gloria Fern{\'a}ndez-Esparrach, Debora Gil, Cristina Rodr{\'\i}guez, and Fernando Vilari{\~n}o.
\newblock Wm-dova maps for accurate polyp highlighting in colonoscopy: Validation vs. saliency maps from physicians.
\newblock {\em Computerized medical imaging and graphics}, 43:99--111, 2015.

\bibitem{jha2020kvasir}
Debesh Jha, Pia~H Smedsrud, Michael~A Riegler, P{\aa}l Halvorsen, Thomas De~Lange, Dag Johansen, and H{\aa}vard~D Johansen.
\newblock Kvasir-seg: A segmented polyp dataset.
\newblock In {\em MultiMedia Modeling: 26th International Conference, MMM 2020, Daejeon, South Korea, January 5--8, 2020, Proceedings, Part II 26}, pages 451--462. Springer, 2020.

\bibitem{codella2019skin}
Noel Codella, Veronica Rotemberg, Philipp Tschandl, M~Emre Celebi, Stephen Dusza, David Gutman, Brian Helba, Aadi Kalloo, Konstantinos Liopyris, Michael Marchetti, et~al.
\newblock Skin lesion analysis toward melanoma detection 2018: A challenge hosted by the international skin imaging collaboration (isic).
\newblock {\em arXiv preprint arXiv:1902.03368}, 2019.

\bibitem{shi2015hierarchical}
Jianping Shi, Qiong Yan, Li~Xu, and Jiaya Jia.
\newblock Hierarchical image saliency detection on extended cssd.
\newblock {\em IEEE transactions on pattern analysis and machine intelligence}, 38(4):717--729, 2015.

\bibitem{wang2017learning}
Lijun Wang, Huchuan Lu, Yifan Wang, Mengyang Feng, Dong Wang, Baocai Yin, and Xiang Ruan.
\newblock Learning to detect salient objects with image-level supervision.
\newblock In {\em Proceedings of the IEEE conference on computer vision and pattern recognition}, pages 136--145, 2017.

\bibitem{wah2011caltech}
Catherine Wah, Steve Branson, Peter Welinder, Pietro Perona, and Serge Belongie.
\newblock The caltech-ucsd birds-200-2011 dataset.
\newblock 2011.

\bibitem{aflalo2023deepcut}
Amit Aflalo, Shai Bagon, Tamar Kashti, and Yonina Eldar.
\newblock Deepcut: Unsupervised segmentation using graph neural networks clustering.
\newblock In {\em Proceedings of the IEEE/CVF International Conference on Computer Vision}, pages 32--41, 2023.

\bibitem{kanungo2002efficient}
Tapas Kanungo, David~M Mount, Nathan~S Netanyahu, Christine~D Piatko, Ruth Silverman, and Angela~Y Wu.
\newblock An efficient k-means clustering algorithm: Analysis and implementation.
\newblock {\em IEEE transactions on pattern analysis and machine intelligence}, 24(7):881--892, 2002.

\end{thebibliography}

\end{document}